\documentclass[journal,twocolumn]{IEEEtran} 
\usepackage[utf8]{inputenc}

\usepackage{amsmath,amssymb,amsthm}
\usepackage{mathptmx} 
\usepackage{amsfonts} 

\usepackage{makecell}

\usepackage{graphicx}
\graphicspath{{figures/}}  
\usepackage{epstopdf}
\usepackage{pgfplots}
\pgfplotsset{compat=1.18}
\usepackage{subcaption}
\usepackage{float}
\usepackage{hyphenat}
\usepackage{changepage} 
\usepackage{placeins}

\usepackage{booktabs}
\usepackage{tabularx,colortbl,multirow,array}

\usepackage{algorithm}
\usepackage{algpseudocode}

\usetikzlibrary{positioning, arrows.meta, shapes.geometric, arrows}

\usepackage[compact]{titlesec}  
\titlespacing*{\section}{0pt}{*1.3}{*0.8}
\titlespacing*{\subsection}{0pt}{*1.2}{*0.68}  

\usepackage{bm}

\usepackage{caption}
\usepackage{hyperref}
\usepackage{enumitem}
\usepackage{tcolorbox}
\tcbuselibrary{fitting}
\usepackage{mdframed}
\usepackage[hang,flushmargin]{footmisc}
\usepackage{cite}
\usepackage{CJKutf8}

\usepackage{url}
\usepackage{xcolor}
\usepackage{amsthm}  
\usepackage{microtype}
\usepackage{etoolbox}
\usepackage{indentfirst} 
\usepackage{geometry}
\usepackage{tikz}
\usetikzlibrary{shapes.geometric, arrows.meta, positioning, calc}

\title{Discovering Sparse Counterfactual Factors via Latent Adjustment for Survey-based Community Intervention}

\author{
    Fatima Ashraf,
    Muhammad Ayub Sabir,
    Junbiao Pang$^*$, Yufang Zhou, and Yan Shang
    \thanks{Email: $^*$ Corresponding author, junbiao\_pang@bjut.edu.cn}%
}

\begin{document}

\maketitle

\begin{abstract}
Transportation surveys are widely used to understand travel preferences and adoption barriers, yet most survey-based analyses remain descriptive or predictive and rarely provide sparse, policy-feasible intervention strategies. We study sparse counterfactual community intervention from survey responses, where the goal is to shift a target respondent group toward a desired reference group through controllable survey-variable adjustments. We formulate this task as a policy-feasible distributional alignment problem using a fixed-basis nonnegative latent representation that preserves pre/post comparability and provides a stable map from latent factors to original variables. To make latent movement actionable, target-relevant latent factors are identified through Shapley-guided attribution and transferred to controllable variables as intervention priorities. Feasible group-level adjustments are then learned by minimizing an entropy-regularized optimal-transport discrepancy between the post-intervention target distribution and the reference distribution, together with a weighted $\ell_{2,1}$ penalty that promotes shared policy-lever sparsity. Experiments on real-world transportation survey datasets show that the proposed framework produces compact and interpretable policy-feasible interventions with explicit adjustment magnitudes, improves population-level conversion, and preserves intervention sparsity. Code and datasets are publicly available at: \url{https://github.com/pangjunbiao/latent-group-alignment.git}.
\end{abstract}



\section{Introduction}

Surveys are widely used to study adoption-related intentions, behavioral barriers, and respondent heterogeneity in public programs, services, and technologies \cite{srivastava2011case},\cite{borhan2014predicting}, \cite{kim2025using}. In transportation settings, for example, large-scale questionnaires help agencies understand who adopts transit, why travel decisions are made, and which controllable factors may shape stated intention under practical constraints \cite{yeganeh2018social}, \cite{jing2022impact}, \cite{susanto2025investigating}. However, most survey-based analyses remain descriptive, explanatory, or predictive: they identify factors associated with adoption intention, but rarely provide a principled mechanism for learning sparse and feasible interventions from survey responses \cite{liu2025incentives},\cite{antipova2020accessibility}, \cite{zhang2024promoting}.

This gap motivates the problem of sparse counterfactual community intervention, where the objective is to shift a target population toward a desired reference population through policy-feasible adjustments on controllable survey variables.

Existing methods address only fragments of this problem. Structural equation modeling and discrete choice models, including SEM and mixed logit, are effective for estimating latent constructs, structural relationships, and heterogeneous behavioral effects \cite{de2023possible, kriswardhana2025uncovering}. However, they typically stop at explanation or factor identification, without producing optimized interventions that specify which controllable variables should be adjusted jointly and by what magnitude. Recent counterfactual and distributional methods move closer to intervention design, but they are primarily developed for instance-level recourse or predictive distribution shifts rather than population-level, survey-based interventions under mixed-type feasibility constraints \cite{wang2025counterfactual, you2024distributional, crupi2024counterfactual}. The key missing capability is therefore an integrated methodology that converts survey evidence into sparse, feasible, and interpretable group-level interventions.

This task is challenging for three coupled reasons. First, survey responses are heterogeneous and mixed-type, making direct group comparisons in the original feature space unreliable. Second, latent factors can reveal coherent respondent structure, but they are not actionable unless their importance can be stably and consistently mapped back to controllable variables. Third, community intervention must move a target group toward a reference group at the distribution level while respecting immutable attributes, mixed-type validity, and sparsity constraints. 

We address these challenges through a conversion-by-alignment framework. It begins by representing encoded survey responses in a fixed-coordinate latent space. We instantiate this representation with fixed-basis NMF because its nonnegative additive factors match the encoded survey domain and its loading matrix provides a stable map from latent factors to original variables. The basis is kept fixed so that pre- and post-intervention respondents are compared under the same latent coordinates.

Within this fixed latent space, outcome-anchored clustering identifies the desired reference group and the target group, and both groups are represented as empirical distributions rather than centroids. To make latent movement actionable, we train a transparent logistic surrogate only as a latent-outcome association probe and use Shapley attribution to obtain target-group latent-factor relevance. These latent priorities are then transferred through the fixed NMF basis to controllable survey variables. Finally, we learn sparse feasible adjustments by minimizing an entropy-regularized OT discrepancy between the post-intervention target distribution and the reference distribution, together with a weighted $\ell_{2,1}$ penalty that promotes shared policy-lever sparsity.

The main contributions of this paper are as follows: 
\begin{itemize}

    \item We formulate survey-based community intervention as a sparse, policy-feasible distributional alignment problem that shifts a target group toward a reference group through controllable survey-variable adjustments in a fixed latent space.

    \item We develop a target-aware latent-to-feature prioritization mechanism, where a transparent logistic surrogate and Shapley attribution identify important latent factors and transfer them through the fixed basis to controllable feature priorities.

    \item We propose a sparse feasible distributional intervention objective that combines entropic OT alignment with weighted $\ell_{2,1}$ shared-lever sparsity to produce compact, interpretable, and policy-feasible adjustments.
\end{itemize}

\section{Related Work}

\subsection{Survey-Based Perception and Intention Analysis in Public Transportation}
Recent studies have examined public-transport perception mainly through analysis of service quality, satisfaction, and behavioral intention. De O\~na and de O\~na \cite{de2023possible} used SEM and SEM-MIMIC on private-vehicle users across five European metropolitan areas and showed that involvement is a key latent factor affecting behavioral intention, while also revealing substantial heterogeneity across user groups. Rong et al. \cite{rong2022impact} combined passenger satisfaction surveys with actual bus traveling-performance data and, using gradient boosting decision trees, showed that passenger perception does not directly mirror objective service performance but follows a complex nonlinear relationship. Ye and Sato \cite{ye2025private} analyzed private-car user's willingness to switch to public transportation through SEM and found that psychological benefits of public transport promote switching intention, whereas favorable perceptions of private cars suppress it. At a broader level, Sogbe et al. \cite{sogbe2025scaling} synthesized 104 studies on bus transport usage in developing countries and identified safety, reliability, comfort, and accessibility as core determinants of public-transport usage, while also highlighting issues such as first-mile/last-mile connectivity and behavioral heterogeneity. Overall, this line of work identifies key perceptual factors and explains differences in public-transport adoption. However, it does not provide a sparse, actionable intervention over controllable variables.

\subsection{Latent-Factor and Topic-Based Representation in Transportation Data}
Recent studies have used topic-based and latent-factor methods to uncover hidden structure in transportation-related data. Ashraf et al. \cite{ashraf2025importance} proposed an importance-aware topic-modeling framework for public-transit risk discovery from noisy social media, showing that low-rank topic structure and topic-localized residual interactions can be jointly recovered for interpretable risk characterization. Li et al. \cite{li2025topic} applied BERTopic to large-scale social-media discourse and showed that topic-based semantic modeling can reveal recurring public concerns, psychological barriers, and service expectations related to urban transit adoption. Yang et al. \cite{yang2023identifying} used NMF on large-scale mobility visitation data to represent urban lifestyles as mixtures of latent patterns not fully explained by demographics. Aminpour and Saidi \cite{aminpour2025unveiling} applied Latent Dirichlet Allocation to transit smart-card and land-use data to infer detailed non-home/work activity patterns, showing that topic models can recover interpretable mobility structure beyond simple trip-purpose categories. Kriswardhana et al. \cite{kriswardhana2025uncovering} combined latent class analysis and structural equation modeling to identify heterogeneous public-transport user profiles and showed that the effects of service quality, satisfaction, involvement, and prior knowledge vary across traveler groups.

These studies show that topic-based and latent-factor models are effective for discovering structure, heterogeneity, and interpretable representations in mobility and transit data. However, they focus on representation and profiling, and do not connect latent representations to actionable intervention over controllable survey factors.

\subsection{Explainable AI and Counterfactual Explanation Methods}

Recent explainable-AI research has increasingly used counterfactual reasoning to generate actionable explanations through structured feature or latent-space adjustments. Na et al. \cite{na2023toward} generated practical and plausible counterfactuals by minimally adjusting semantic information in a disentangled latent space, while Crupi et al. \cite{crupi2024counterfactual} proposed latent-space interventions that improve the feasibility and causal consistency of counterfactual recommendations. Kim et al. \cite{kim2024cirf} further showed that incorporating dependencies among related features can improve the plausibility and coherence of counterfactual changes beyond feature-independent perturbations. In the transportation domain, Wang et al. \cite{wang2025counterfactual} combined deep traffic forecasting with multi-objective counterfactual optimization to explain how contextual variables affect traffic-speed predictions under spatial, temporal, and scenario constraints.

Overall, these studies show that counterfactual reasoning can move beyond feature attribution toward feasible and interpretable explanation mechanisms. However, they are primarily designed for instance-level recourse or model explainability, rather than survey-based group conversion with controllable policy levers and sparse feasible intervention design.

\section{Methodology}
\subsection{Policy-Feasible Survey Intervention Problem}
\label{subsec:data_partition}

We consider a survey dataset comprising $n$ respondents and $d$ encoded features. Each respondent $i$ is represented by a nonnegative vector $x_i \in \mathbb{R}_+^d$, and the full survey matrix is $X = [x_1^\top, \dots, x_n^\top]^\top \in \mathbb{R}_+^{n \times d}$. The nonnegative representation is used not only as preprocessing, but also to provide a common domain for ordinal, one-hot, and numeric survey variables while preserving feature-level interpretability.

This work differs from standard survey modeling and instance-level counterfactual explanation. Rather than only identifying variables associated with an outcome or generating isolated individual recourse \cite{verma2024counterfactual}, we seek sparse and feasible changes to controllable survey variables that move a low-outcome target group toward a desired reference group at the population level. Since survey variables have different measurement semantics, intervention feasibility must be defined before any latent modeling or optimization is introduced.

To structure the heterogeneous survey responses, the encoded features are first partitioned by measurement type:
\begin{equation}
\label{eq:type_partition_method}
\{1, \dots, d\} = S_{\mathrm{Lik}} \;\dot\cup\; S_{\mathrm{Cat}} \;\dot\cup\; S_{\mathrm{Num}},
\end{equation}
where $S_{\mathrm{Lik}}$, $S_{\mathrm{Cat}}$, and $S_{\mathrm{Num}}$ denote Likert, categorical one-hot, and numeric variables, respectively. Binary variables are treated as a special subset $S_{\mathrm{Bin}} \subseteq S_{\mathrm{Num}}$. This measurement-type partitioning allows the framework to handle mixed-type survey data appropriately, enforcing type-specific feasibility constraints during interventions.

To distinguish actionable levers from immutable attributes, we further partition the feature indices as
\begin{equation}
\label{eq:ctrl_fixed_partition_method}
\{1, \dots, d\} = S_{\mathrm{ctrl}} \;\dot\cup\; S_{\mathrm{fixed}}, \qquad S_{\mathrm{ctrl}} \cap S_{\mathrm{fixed}} = \emptyset,
\end{equation}
where $S_{\mathrm{ctrl}}$ indexes controllable factors and $S_{\mathrm{fixed}}$ indexes immutable attributes. This distinction is critical: not all influential variables can be directly intervened upon, and policy feasibility requires that interventions only modify controllable features.

\textbf{Problem.} Let $\Delta_i \in \mathbb{R}^d$ denote the intervention applied to respondent $i$, and let $\Delta = [\Delta_1^\top, \dots, \Delta_n^\top]^\top \in \mathbb{R}^{n \times d}$ denote the full intervention matrix. The post-intervention feature vector for respondent $i$ is $x_i^{\mathrm{post}}=x_i+\Delta_i$. Given a target group $B$ and a desired reference group $A$, our goal is to learn a sparse, feasible intervention $\Delta$ over $S_{\mathrm{ctrl}}$ that shifts the post-intervention target distribution toward the reference distribution, while leaving non-target respondents and the reference group unchanged. The intervention must respect the following feasibility constraints: 

\begin{enumerate}[label=(\roman*)]
    \item \textbf{Immutability:} $\Delta_{ij} = 0$ for all $i$ and $j \in S_{\mathrm{fixed}}$.
    \item \textbf{Mixed-type validity:} $x_i^{\mathrm{post}}$ must remain valid under the encoded survey representation. Specifically, Likert variables must remain within admissible ordinal ranges, numeric variables within feasible intervals, categorical one-hot blocks are not directly modified, and binary controllable variables are relaxed to $[0,1]$ during optimization and rounded to $\{0,1\}$ when reporting interventions.
\end{enumerate}

Thus, the intervention problem is not defined by predictive relevance alone, but by the joint requirements of actionability, mixed-type validity, group-level distributional movement, and sparse policy implementation.

\subsection{Fixed-Coordinate Latent Representation}
\label{subsec:nmf}

The encoded survey space is heterogeneous and often high-dimensional, making direct group comparison in the original feature space difficult to interpret. We therefore evaluate group structure and intervention effects in a lower-dimensional latent space. However, not every latent representation is suitable for distributional counterfactual intervention. In our setting, the latent space must satisfy three requirements: compatibility with nonnegative encoded survey responses, an interpretable link between latent factors and original variables, and a fixed coordinate system in which pre- and post-intervention respondents remain comparable.

We employ standard NMF as a representational tool to construct this latent space \cite{saberi2025nonnegative}. NMF is used because its nonnegative additive factors are compatible with the encoded survey representation, and its explicit loading matrix provides a stable map from latent coordinates back to original variables.

Other latent models could be used only if they keep the same fixed coordinates and provide such a stable latent-to-feature map. Unlike signed linear bases or nonlinear embeddings, NMF offers a simple and interpretable way to satisfy these requirements for nonnegative survey responses.

Specifically, the encoded survey matrix \(X\) is factorized as
\begin{equation}
\label{eq:nmf_factorization}
X \approx WH,
\qquad
W \in \mathbb{R}_+^{n\times k},
\quad
H \in \mathbb{R}_+^{k\times d},
\quad
k \ll \min(n,d),
\end{equation}
where the \(i\)-th row \(w_i^\top\) of \(W\) is the latent representation of respondent \(i\), and the \(r\)-th row of \(H\) describes the loading pattern of latent factor \(r\) over the encoded features.

The key design choice is to keep the basis \(H\) fixed after it is learned from the observed survey data. This fixed-coordinate design is necessary because the intervention changes \(x_i\), but the meaning of each latent factor must remain unchanged. If the basis were re-learned after intervention, a change in latent coordinates could reflect either a real respondent movement or a change in the coordinate system itself, making pre/post comparison ambiguous. By fixing \(H\), every observed and post-intervention respondent is represented using the same latent factors, so movement in latent space has a consistent interpretation.

Because NMF is scale-ambiguous, we normalize each row of \(H\) to unit \(\ell_1\) norm and absorb scale into \(W\). Since group identification and optimal-transport alignment depend on relative latent composition, we use row-normalized coefficients \(\tilde{w}_i=w_i/\|w_i\|_1\).

For any post-intervention feature vector \(x_i^{\mathrm{post}}=x_i+\Delta_i\), its latent representation is obtained by projection onto the fixed basis via nonnegative least squares (NNLS):
\begin{equation}
\label{eq:post_nnls_method}
w_i^{\mathrm{post}}
=
\arg\min_{w\ge 0}
\|x_i^{\mathrm{post}} - wH\|_2^2.
\end{equation}

Thus, fixing \(H\) does not freeze respondent positions in latent space; post-intervention respondents can still move through their updated coefficients \(w_i^{\mathrm{post}}\), while the basis remains a common reference frame.

The normalized post-intervention coefficient is \(\tilde{w}_i^{\mathrm{post}}=w_i^{\mathrm{post}}/\|w_i^{\mathrm{post}}\|_1\). Thus, interventions are applied in the original feature space, while effects are measured as stable, interpretable movement from \(\tilde{w}_i\) to \(\tilde{w}_i^{\mathrm{post}}\) in the shared fixed latent space. This fixed-coordinate representation provides the common geometry used for group identification, latent-to-feature prioritization, and distributional alignment.

\subsection{Outcome-Anchored Latent Group Distributions}
\label{subsec:clustering}

Community intervention requires shifting a target group toward a reference group; therefore, we first identify coherent respondent groups in the shared latent space. The normalized latent representations $\tilde{w}_i$ from Sec.~\ref{subsec:nmf} encode relative latent composition and provide a fixed-coordinate geometry for group comparison.

We apply $k$-means clustering to $\{\tilde{w}_i\}_{i=1}^n$ with $G$ clusters, yielding assignments $g(i)\in\{1,\dots,G\}$ and cluster index sets $\mathcal{I}_c=\{i:\,g(i)=c\}$. These clusters capture latent response patterns, but latent structure alone does not determine which group should serve as the intervention target or the desired reference group. To resolve this, we anchor the clusters to the observed survey outcome. Let $y_i$ denote the primary outcome score, and define the cluster-wise mean

\begin{equation}
\label{eq:cluster_mean_intention}
\bar{y}(c)=\frac{1}{|\mathcal{I}_c|}\sum_{i\in\mathcal{I}_c}y_i.
\end{equation}

The cluster with the highest mean outcome is selected as the reference group, $A=\arg\max_c \bar{y}(c)$, while the cluster with the lowest mean outcome is selected as the target group, $B=\arg\min_c \bar{y}(c)$. Their respondent index sets are denoted by $\mathcal{I}_A$ and $\mathcal{I}_B$, respectively. This outcome anchoring separates latent representation from intervention role assignment: clustering identifies coherent respondent structure, while the outcome determines which latent group represents the desired direction and which group defines the intervention target.

Having identified the target and reference groups, we represent them as empirical distributions rather than single centroids. This distributional representation is essential because the intervention objective aims to shift a population distribution while preserving within-group heterogeneity, which cannot be captured by a single mean vector. Formally, we define
\begin{equation}
\label{eq:empirical_measures}
\mu_A=\frac{1}{|\mathcal{I}_A|}\sum_{j\in\mathcal{I}_A}\delta_{\tilde{w}_j},
\qquad
\mu_B=\frac{1}{|\mathcal{I}_B|}\sum_{i\in\mathcal{I}_B}\delta_{\tilde{w}_i},
\end{equation}
where $\delta_w$ denotes the Dirac mass at location $w$. These empirical measures define the group distributions used later for optimal-transport alignment. Unlike centroid-based representations, they preserve the variability and heterogeneity of respondents within each group.

\subsection{Target-Aware Latent-to-Feature Prioritization}
\label{subsec:shapley}

The target and reference distributions define the desired population shift, but they do not specify which controllable survey variables should be prioritized for intervention. This step is critical because the group structure exists in latent space, while feasible policy actions must operate directly on the original encoded survey variables. We therefore need a bridge from outcome-relevant latent factors to actionable feature-level priorities.

To identify outcome-relevant latent factors, we use a transparent latent-outcome probe. Specifically, we train a surrogate model on normalized latent coefficients $\tilde{w}_i$ to predict a binarized outcome label $y_i^{\mathrm{bin}} \in \{0,1\}$ derived from $y_i$ (see Supplementary Sec.~S1):
\begin{equation}
\label{eq:surrogate_model}
\hat{y}_i = f(\tilde{w}_i),
\qquad
f:\mathbb{R}_+^k \to [0,1].
\end{equation}
We use logistic regression for $f$ because the purpose of the surrogate is not to build a high-capacity predictor, but to obtain a stable and transparent association between the fixed latent coordinates and the survey outcome. Since each input dimension of \(f\) corresponds to a fixed latent factor, Shapley attribution can be interpreted directly as latent-factor relevance \cite{borgonovo2024many}. The surrogate is therefore not treated as a causal model or the final intervention objective; it serves only as an outcome-association probe for prioritizing latent factors before mapping them back to controllable variables.

For each respondent $i$ and latent factor $r \in \{1,\dots,k\}$, let $\phi_r^{(i)}$ denote the local Shapley value of factor $r$ for the prediction $f(\tilde{w}_i)$. Since interventions target the low-outcome group, we aggregate local attributions over $\mathcal{I}_B$ to obtain target-group-specific latent-factor relevance:
\begin{equation}
\label{eq:target_group_shapley}
\varphi_r
=
\frac{1}{|\mathcal{I}_B|}
\sum_{i\in\mathcal{I}_B}
|\phi_r^{(i)}|.
\end{equation}
The absolute value ensures that $\varphi_r$ reflects the magnitude of predictive relevance irrespective of direction, focusing on feature importance rather than effect sign. The top-$q$ most relevant latent factors form the set:
\begin{equation}
\label{eq:topq_factors}
\mathcal{K}_q
=
\operatorname{Top}\text{-}q\big(\{1,\dots,k\};\,\varphi_r\big).
\end{equation}

These selected latent factors are then mapped to actionable feature priorities using the fixed NMF basis $H$ from Sec.~\ref{subsec:nmf}, where $H_{rj}$ denotes the loading of feature $j$ on latent factor $r$. Each controllable feature $j \in S_{\mathrm{ctrl}}$ receives a priority score:
\begin{equation}
\label{eq:feature_weight_from_factors}
\omega_j
=
\sum_{r \in \mathcal{K}_q} \varphi_r\, H_{rj}.
\end{equation}

A large $\omega_j$ indicates that controllable feature $j$ has high loading, through the fixed basis $H$, on latent factors relevant to the target-group outcome. Thus, $\omega_j$ is not a causal effect estimate; it is a target-aware priority used to guide sparse intervention optimization. This latent-to-feature transfer is possible because the latent coordinates and feature loadings remain fixed across observed and post-intervention respondents.

\subsection{Sparse Feasible Distributional Intervention}
\label{subsec:intervention}

Given the target and reference distributions $(\mu_B,\mu_A)$ from Sec.~\ref{subsec:clustering} and the controllable-feature priorities $\{\omega_j\}_{j\in S_{\mathrm{ctrl}}}$ from Sec.~\ref{subsec:shapley}, the final step is to learn an intervention that is feasible in the original survey-feature space and effective in the fixed latent space. Let $\Delta\in\mathbb{R}^{n\times d}$ denote the intervention matrix, with post-intervention response vector $x_i^{\mathrm{post}}=x_i+\Delta_i$. Interventions are applied only to the target group and must satisfy the actionability and mixed-type validity constraints defined in Sec.~\ref{subsec:data_partition}. We denote the corresponding feasible set by $C$:
\begin{equation}
\label{eq:feasible_set_C_main}
\begin{aligned}
C=\Big\{\Delta \in \mathbb{R}^{n\times d}:\;&
\Delta_i=0,\ \forall i\notin\mathcal{I}_B,\\
&\Delta_{ij}=0,\ \forall j\in S_{\mathrm{fixed}}\cup S_{\mathrm{Cat}},\\
&x_i+\Delta_i \text{ remains feasible},\ \forall i\in\mathcal{I}_B
\Big\}.
\end{aligned}
\end{equation}

For any $\Delta\in C$, each post-intervention target respondent is projected onto the fixed basis $H$ using Eq.~\eqref{eq:post_nnls_method}, yielding normalized latent code $\tilde{w}_i^{\mathrm{post}}$. The resulting post-intervention target distribution is
\begin{equation}
\label{eq:muB_post}
\mu_B^{\mathrm{post}}(\Delta)
=
\frac{1}{|\mathcal{I}_B|}
\sum_{i\in\mathcal{I}_B}\delta_{\tilde{w}_i^{\mathrm{post}}}.
\end{equation}
Thus, the intervention is optimized in the original feature space, but its population-level effect is evaluated as movement of $\mu_B^{\mathrm{post}}(\Delta)$ toward $\mu_A$ in the fixed latent space.

Because both groups are represented as empirical distributions in the fixed latent space, the alignment criterion must compare full distributions rather than only individual points or group means. We therefore use entropic optimal transport to measure the minimum latent-space transport cost required to move the post-intervention target distribution toward the reference distribution \cite{peyre2019computational}.

Let $\mathcal{I}_A=\{a_q\}_{q=1}^{n_A}$ and $\mathcal{I}_B=\{b_p\}_{p=1}^{n_B}$. The intervention-dependent ground cost is
\begin{equation}
\label{eq:cost_matrix_post}
M_{pq}(\Delta)=\|\tilde{w}_{b_p}^{\mathrm{post}}-\tilde{w}_{a_q}\|_2^2.
\end{equation}

This ground cost is computed in the fixed latent space, so distances measure respondent similarity under a shared coordinate system.

\begin{figure}[t]
\centering
\includegraphics[width=0.48\textwidth]{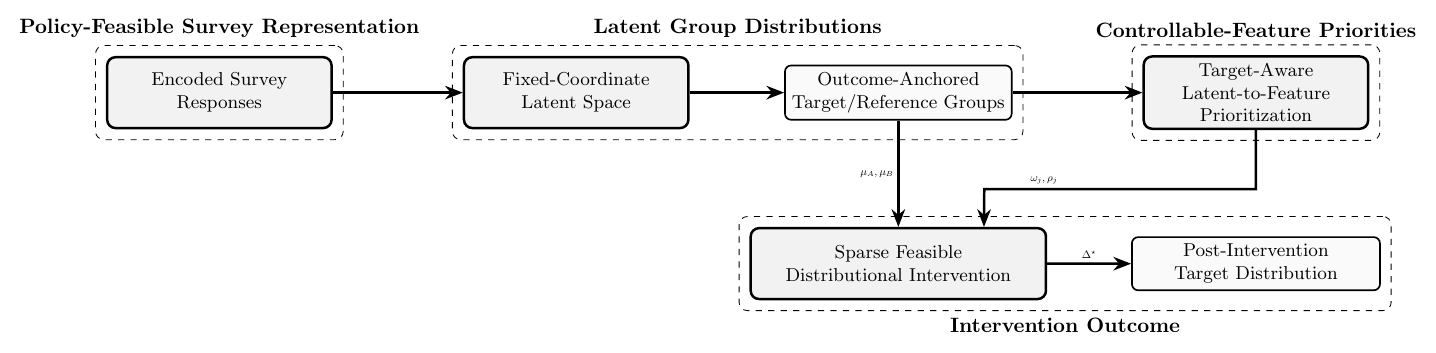}
\caption{Overview of the conversion-by-alignment framework for survey-based community intervention.}
\label{fig:method_overview}
\end{figure}

The entropic OT discrepancy is
\begin{equation}
\label{eq:ot_entropic_post}
W_{\eta}(\mu_B^{\mathrm{post}}(\Delta),\mu_A)
=
\min_{\Gamma\in\mathbb{R}_+^{n_B\times n_A}}
\langle \Gamma,M(\Delta)\rangle
+
\eta\sum_{p,q}\Gamma_{pq}(\log\Gamma_{pq}-1),
\end{equation}
subject to
\[
\Gamma\mathbf{1}=\frac{1}{n_B}\mathbf{1}, \qquad 
\Gamma^\top\mathbf{1}=\frac{1}{n_A}\mathbf{1},
\]
where $\Gamma$ is the transport plan and $\eta>0$ controls entropic smoothing. Unlike centroid matching, this objective aligns empirical distributions and therefore accounts for within-group heterogeneity. The entropy term provides a smoother and more stable alignment objective for optimization.

Distributional alignment alone may spread changes across many variables. To obtain policy-realistic interventions, we impose weighted shared-lever sparsity:
\begin{equation}
\label{eq:weighted_l21_main}
\Omega(\Delta;\rho)
=
\sum_{j\in S_{\mathrm{ctrl}}}\rho_j\|\Delta_{:,j}\|_2,
\qquad
\rho_j=(\omega_j+\varepsilon_\omega)^{-1}.
\end{equation}
Here $\|\Delta_{:,j}\|_2$ measures the population-level adjustment magnitude of controllable feature $j$. The $\ell_{2,1}$ structure encourages the intervention to activate a compact set of shared policy levers, while the weights $\rho_j$ reduce the penalty on target-relevant features identified in Sec.~\ref{subsec:shapley}.

The final intervention is obtained by solving
\begin{equation}
\label{eq:conversion_optimization_main}
\min_{\Delta\in C}
\;
W_{\eta}\big(\mu_B^{\mathrm{post}}(\Delta),\mu_A\big)
+
\lambda\,\Omega(\Delta;\rho),
\end{equation}
where $\lambda>0$ balances distributional alignment and sparse policy implementation. This objective combines the central requirements of the framework: feasible feature-space adjustment, fixed-coordinate latent evaluation, distribution-level movement toward the reference group, and sparse shared policy levers.

The optimization is nonconvex because $\Delta$ affects the latent distribution through the fixed-basis projection in Eq.~\eqref{eq:post_nnls_method}. We solve it using an auxiliary-variable penalized relaxation with post-intervention latent codes $U$ and alternating projected proximal updates. Projection onto $C$ maintains feasibility at each update. Detailed update rules and convergence criteria are provided in the Supplementary Material (Sec.~I). An overview of the proposed framework is shown in Fig.~\ref{fig:method_overview}, and the overall procedure is summarized in Algorithm~\ref{alg:workflow_final}.

\begin{algorithm}[H]
\caption{Conversion-by-Alignment Framework}
\label{alg:workflow_final}
\begin{algorithmic}[1]
\Require Encoded feature matrix $X\in\mathbb{R}_+^{n\times d}$; controllable/fixed partitions $S_{\mathrm{ctrl}},S_{\mathrm{fixed}}$; outcome scores $\{y_i\}_{i=1}^n$; NMF rank $k$; cluster number $G$; OT regularization $\eta$; sparsity weight $\lambda$
\Ensure Sparse counterfactual intervention $\Delta^\star$ and active policy levers

\State Factorize $X\approx WH$ using \eqref{eq:nmf_factorization}; fix $H$ and obtain row-normalized latent representations $\tilde{w}_i$.
\State Cluster $\{\tilde{w}_i\}_{i=1}^n$ into $G$ groups; define reference group $A$ and target group $B$ using \eqref{eq:cluster_mean_intention}.
\State Construct empirical group distributions $\mu_A$ and $\mu_B$ using \eqref{eq:empirical_measures}.
\State Train the latent-outcome surrogate $f$ using \eqref{eq:surrogate_model}; compute local Shapley values $\phi_r^{(i)}$, aggregate $\varphi_r$ using \eqref{eq:target_group_shapley}, select $\mathcal{K}_q$ using \eqref{eq:topq_factors}, and compute feature priorities $\omega_j$ using \eqref{eq:feature_weight_from_factors}.
\State Define feasible set $C$ using \eqref{eq:feasible_set_C_main}; initialize $\Delta^{(0)}=0$ with support restricted to $\mathcal{I}_B\times S_{\mathrm{ctrl}}$.
\State Initialize auxiliary latent codes $U^{(0)}\in\mathbb{R}_+^{n_B\times k}$ and set $\rho_j=(\omega_j+\varepsilon_\omega)^{-1}$ for $j\in S_{\mathrm{ctrl}}$.

\For{$t=0,1,\dots$ until convergence}
    \State Update latent codes $U^{(t+1)}$ via projected gradient step (Supp. Sec.~I).
    \State Update intervention $\Delta_B^{(t+1)}$ via projected proximal-gradient step with weighted $\ell_{2,1}$ regularization (Supp. Sec.~I).
    \State Project onto $C$ and evaluate $\mathcal{J}(\Delta^{(t+1)},U^{(t+1)})$ for convergence.
\EndFor

\State Set $\Delta^\star$ to the final feasible intervention and return active levers $\{j:\|\Delta^\star_{:,j}\|_2>0\}$.
\end{algorithmic}
\end{algorithm}

\section{Results and Discussion}
\subsection{Dataset and Evaluation Metrics}
\label{subsec:evaluation}

\subsubsection{Dataset}

The primary dataset is a real-world questionnaire survey conducted in Beijing to study public-transport adoption, stated behavioral intention, and responses to carbon-incentive mechanisms. It contains 1021 respondents and 40 raw survey fields. After excluding two non-analytic fields (respondent identifier and contact information), the dataset retains 38 substantive variables spanning attitudes, incentive preferences, reward-response behavior, and respondent background attributes (Table~\ref{tab:dataset_summary_combined}). To examine generality beyond this setting, we also consider MNIST as a controlled toy domain and the VTA 2013 on-board transit survey~\cite{vta2013} as a second real-world survey dataset.

\begin{table}[H]
\centering
\caption{Summary of the Beijing and VTA survey datasets.}
\label{tab:dataset_summary_combined}
\scriptsize
\setlength{\tabcolsep}{3pt}
\renewcommand{\arraystretch}{0.9}
\begin{tabular}{l c}
\toprule
\multicolumn{2}{c}{\textbf{Beijing Survey}} \\
\midrule
Survey respondents & 1021 \\
Raw survey fields & 40 \\
Excluded non-analytic fields & 2 \\
Substantive variables & 38 \\
Survey domain & Public transport carbon-incentive behavior \\
Variable types & Attitudinal / Incentive / Behavioral / Demographic \\
\midrule
\multicolumn{2}{c}{\textbf{VTA Survey}} \\
\midrule
Survey records & 9654 \\
Raw survey fields & 85 \\
Survey domain & On-board transit behavior and service evaluation \\
Variable types & Trip / Access-Transfer / Fare / Rating / Demographic \\
\bottomrule
\end{tabular}
\end{table}

At the raw-variable level, the Beijing survey consists of four parts: (i) attitudinal and intention-related items, (ii) policy and incentive items, (iii) reward-response items under different incentive levels, and (iv) demographic and contextual attributes such as occupation, education, income, vehicle ownership, residential district, destination district, and transit-card usage. This structure is well suited to latent-factor modeling because the observed survey responses are heterogeneous but partially driven by shared underlying preferences. In particular, attitudinal items, incentive preferences, and reward-response variables can be interpreted as observable manifestations of a smaller set of adoption-related latent factors, while demographic and contextual attributes provide background information that remains fixed under intervention.

The VTA dataset is a real-world on-board transit survey conducted in 2013 to study rider travel behavior, service evaluation, and transit-use characteristics. The dataset contains 9,654 survey records and 85 raw fields, covering trip origins and destinations, access and egress modes, transfer behavior, fare payment, service-quality ratings, transit-use frequency, demographic attributes, language background, vehicle availability, and stated alternatives if VTA transit were unavailable. This structure provides both service-related variables that can be treated as potentially actionable factors and respondent-specific background attributes that remain fixed, making the dataset suitable for evaluating group-level behavioral shift in a second real-world survey domain.

\subsubsection{Evaluation Metrics}

The proposed framework aims to learn a minimal, policy-feasible intervention that moves the target group toward the reference (green-accepting) group. Accordingly, we evaluate both intervention effectiveness and intervention cost.

We use respondent-level conversion as the primary outcome. Let $\mathcal{I}_B$ denote the target group and let $n_B = |\mathcal{I}_B|$. Using the surrogate score $f(\cdot)$ on normalized latent representations, a target-group respondent is counted as converted if the predicted score crosses a decision threshold $\tau_y$ after intervention. The number of converted respondents is
\begin{equation}
N_{\mathrm{conv}} =
\sum_{i\in\mathcal I_B} 
\mathbf{1}\{ f(\tilde w_i) < \tau_y \;\text{and}\; f(\tilde w_i^{\mathrm{post}}) \ge \tau_y \},
\end{equation}
and the corresponding conversion rate is $R_{\mathrm{conv}} = N_{\mathrm{conv}} / n_B$. In our experiments, we use $\tau_y = 0.5$, which is the standard decision threshold for the logistic-regression surrogate.

Intervention effort is measured by the group-sparse magnitude of the learned adjustment:
\begin{equation}
\|\Delta^\star\|_{2,1} = \sum_{j\in S_{\mathrm{ctrl}}} \|\Delta^\star_{:,j}\|_2,
\end{equation}
which quantifies the total change applied across controllable policy levers. Policy sparsity is further characterized by the number of activated levers:
\begin{equation}
N_{\mathrm{lever}}(\tau_\Delta) = 
\sum_{j\in S_{\mathrm{ctrl}}} \mathbf{1}\{\|\Delta^\star_{:,j}\|_2 > \tau_\Delta\},
\end{equation}
where $\tau_\Delta$ is a small numerical tolerance used to distinguish active levers from negligible changes. In our experiments, we set $\tau_\Delta = 10^{-6}$.

To measure intervention efficiency, we define
\begin{equation}
\mathrm{Eff}_{\mathrm{conv}} = \frac{N_{\mathrm{conv}}}{\|\Delta^\star\|_{2,1}},
\end{equation}
which reports the number of converted respondents per unit intervention effort.

To evaluate latent-space alignment, we report the reduction in entropic OT discrepancy between the target and reference distributions. Let
\begin{equation}
W_\eta^{\mathrm{before}} = W_\eta(\mu_B,\mu_A),
\qquad
W_\eta^{\mathrm{after}} = W_\eta(\mu_B^{\mathrm{post}}(\Delta^\star), \mu_A).
\end{equation}
We measure alignment improvement by the absolute reduction $\Delta W_\eta = W_\eta^{\mathrm{before}} - W_\eta^{\mathrm{after}}$, and by the relative reduction $\rho_\eta = \frac{\Delta W_\eta}{W_\eta^{\mathrm{before}}}$. These provide complementary views of how effectively the learned intervention aligns the target distribution with the reference distribution.

\subsection{Model Performance on the Survey Dataset}

The proposed framework demonstrates strong effectiveness in moving the target group toward the green-accepting reference group. In the main reference run, the intervention yields 61 conversions out of 355 target respondents, together with a consistent increase in predicted adoption probability and a clear reduction in latent-space discrepancy between the two groups. This indicates that the learned intervention does not merely improve individual surrogate scores, but also moves the overall target distribution closer to the reference group in the learned latent representation. At the same time, the intervention remains sparse and interpretable. Only 8 policy levers are activated, indicating that the observed shift is achieved through focused adjustments rather than broad changes across variables. 

Across multiple random seeds (42-45), the same overall pattern is observed: the intervention consistently improves conversion and alignment while maintaining a sparse structure, as shown in Table \ref{main model results}.

\begin{table}[H]
\centering
\caption{Performance of the proposed method on the survey dataset.}
\label{main model results}
\resizebox{\columnwidth}{!}{
\begin{tabular}{lccccccc}
\toprule
Setting & 
$N_{\mathrm{conv}} \uparrow$ & 
$R_{\mathrm{conv}} \uparrow$ & 
Mean $\Delta p \uparrow$ & 
$N_{\mathrm{lever}} \downarrow$ & 
Effort $(\|\Delta\|_{2,1}) \downarrow$ & 
$\Delta W_\eta \uparrow$ & 
$\rho_\eta \uparrow$ \\
\midrule
Seed 42 & 61 & 0.1718 & 0.0365 & 8 & 15.6777 & 0.1090 & 0.2425 \\
Mean $\pm$ Std & 40.25 $\pm$ 15.42 & 0.1233 $\pm$ 0.0344 & 0.0239 $\pm$ 0.0088 & 8.00 $\pm$ 0.00 & 14.8455 $\pm$ 0.6057 & -- & 0.2122 $\pm$ 0.0367 \\
\bottomrule
\end{tabular}
}
\end{table}

\subsection{Comparison with Baselines and Ablations}

\paragraph{Baselines.}
\begin{itemize}
    \item \textbf{Top-Shapley Single-Lever:} This intervenes on only one controllable variable, namely the lever with the highest Shapley importance. It represents the simplest importance-based intervention rule and tests whether acting on only the single most influential lever is sufficient to induce meaningful group-level movement.

    \item \textbf{Top-Shapley Top-$k$ Uniform:} This selects the top-$k$ controllable levers according to Shapley importance and applies the same uniform intervention to all of them. It tests whether importance ranking alone is enough, without solving the structured optimization problem used in the full method.

    \item \textbf{Max-Coverage Top-$k$ Uniform:} This baseline selects the $k$ levers with the broadest feasible coverage across the target group and applies a uniform intervention over them. It tests whether choosing widely applicable levers is more effective than choosing importance-ranked levers.

    \item \textbf{Outcome-Only Sparse:} This baseline performs a sparse intervention guided only by the direct outcome objective, without the optimal-transport alignment term. It tests whether improving the surrogate outcome alone is sufficient, or whether explicit distributional movement toward the reference group is necessary.
\end{itemize}

Table~\ref{tab:baseline_comparison} shows that simple ranking-based or coverage-based intervention rules produce only limited improvement. Although some baselines achieve lower effort or activate fewer levers, they remain clearly below the full method in conversion, mean probability lift, and latent alignment improvement. This indicates that identifying important levers alone is not sufficient; the intervention must also be designed to move the target group toward the reference group in the learned latent space.

\begin{table}[H]
\centering
\caption{Baselines comparison on survey dataset}
\label{tab:baseline_comparison}
\scriptsize
\setlength{\tabcolsep}{2pt}
\resizebox{\columnwidth}{!}{
\begin{tabular}{lccccccc}
\toprule
Method & 
$N_{\mathrm{conv}} \uparrow$ & 
$R_{\mathrm{conv}} \uparrow$ & 
Mean $\Delta p \uparrow$ & 
Effort $(\|\Delta^*\|_{2,1}) \downarrow$ & 
$N_{\mathrm{lever}}(\tau) \downarrow$ & 
$\Delta W_{\eta} \uparrow$ & 
$\rho_{\eta} \uparrow$ \\
\midrule
\textbf{Full method} & \textbf{61} & \textbf{0.1718} & \textbf{0.0365} & 15.6777 & 8 & \textbf{0.1090} & \textbf{0.2425} \\
Top-Shapley Single-Lever & 14 & 0.0394 & 0.0096 & \textbf{3.7683} & \textbf{1} & 0.0296 & 0.0657 \\
Top-Shapley Top-$k$ Uniform & 36 & 0.1014 & 0.0214 & 9.4207 & 5 & 0.0641 & 0.1444 \\
Max-Coverage Top-$k$ Uniform & 33 & 0.0930 & 0.0194 & 9.4207 & 5 & 0.0579 & 0.1304 \\
Outcome-Only Sparse & 9 & 0.0254 & 0.0045 & 9.4207 & 5 & 0.0143 & 0.0286 \\
\bottomrule
\end{tabular}
}
\end{table}

\paragraph{Ablations.}
\begin{itemize}
    \item \textbf{w/o Shapley weighting:} Removes the Shapley-guided lever weighting and tests whether transferring latent importance back to actionable policy levers is necessary.

    \item \textbf{w/o sparsity:} Removes the sparsity-inducing regularization and tests whether the method can remain effective while preserving a compact and interpretable intervention set.

    \item \textbf{w/o OT alignment:} Removes the optimal-transport alignment objective and tests whether distribution-level matching is necessary beyond simpler alignment criteria.

\end{itemize}

Table~\ref{tab:ablation_comparison} shows that the full method depends on the interaction of all major components. Removing Shapley weighting leads to the most severe degradation, indicating that principled prioritization of controllable levers is central to the method. Removing sparsity preserves much of the raw conversion performance, but does so at the cost of activating nearly all controllable levers, which substantially weakens interpretability and policy realism. Removing OT alignment also slightly reduces conversion and alignment performance, confirming that explicit distributional matching contributes beyond optimizing the surrogate outcome alone.
\begin{table}[H]
\centering
\caption{Ablation study of the proposed method.}
\label{tab:ablation_comparison}
\scriptsize
\setlength{\tabcolsep}{2pt}
\resizebox{\columnwidth}{!}{
\begin{tabular}{lccccccc}
\toprule
Method & 
$N_{\mathrm{conv}} \uparrow$ & 
$R_{\mathrm{conv}} \uparrow$ & 
Mean $\Delta p \uparrow$ & 
Effort $(\|\Delta^*\|_{2,1}) \downarrow$ & 
$N_{\mathrm{lever}}(\tau) \downarrow$ & 
$\Delta W_{\eta} \uparrow$ & 
$\rho_{\eta} \uparrow$ \\
\midrule
\textbf{Full method} & \textbf{61} & \textbf{0.1718} & \textbf{0.0365} & 15.6777 & 8 & \textbf{0.1090} & \textbf{0.2425} \\
w/o Shapley weighting & 8 & 0.0225 & 0.0041 & \textbf{2.1102} & \textbf{7} & 0.0127 & 0.0281 \\
w/o sparsity & 59 & 0.1662 & 0.0358 & 19.3677 & 32 & 0.1069 & 0.2375 \\
w/o OT alignment & 59 & 0.1662 & 0.0356 & 19.3677 & 32 & 0.1059 & 0.2357 \\
\bottomrule
\end{tabular}
}
\end{table}

\subsection{Sensitivity Analysis}

To assess the robustness of the proposed framework, we conduct a targeted sensitivity analysis over three key hyperparameters: the latent dimension ($k$), the number of clusters ($G$), and the sparsity weight ($\lambda_{2,1}$). These parameters control, respectively, (i) the capacity of the latent representation, (ii) the granularity of group structure, and (iii) the strength of sparsity regularization.

The latent dimension $k$ controls how much structure can be captured in the learned latent representation. When $k$ is too small, the representation is too limited to separate meaningful patterns, resulting in negligible intervention effects. As shown in Table~\ref{tab:sensitivity_analysis}, increasing $k$ improves performance, with $k=10$ giving the strongest overall results among the tested settings.

\begin{table}[H]
\centering
\caption{Sensitivity analysis over latent dimension ($k$), number of clusters ($G$), and sparsity weight ($\lambda_{2,1}$).}
\label{tab:sensitivity_analysis}
\vspace{-2mm}
\tiny
\setlength{\tabcolsep}{1pt}
\renewcommand{\arraystretch}{0.82}
\resizebox{\columnwidth}{!}{
\begin{tabular}{l c c c c c c c c c}
\toprule
Parameter & Value & \makecell{$N_{\mathrm{conv}}$} & \makecell{$R_{\mathrm{conv}}$} & \makecell{Mean\\$\Delta p$} & \makecell{$N_{\mathrm{lever}}$\\$(\tau)$} & \makecell{Effort\\$(\|\Delta\|_{2,1})$} & \makecell{$\rho_{\eta}$} & \makecell{Best\\Total Obj.} & Status \\
\midrule
\multirow{4}{*}{\textbf{Latent Dim ($k$)}} 
& 4  & 0  & 0.0000 & 0.0010 & 9  & 1.296  & 0.0009 & -- & Weak \\
& 6  & 0  & 0.0000 & 0.0150 & 5  & 9.631  & 0.0347 & -- & Weak \\
& 8  & 18 & 0.0650 & 0.0168 & 6  & 9.887  & 0.1075 & -- & Moderate \\
& 10 & 61 & 0.1718 & 0.0365 & 8  & 15.678 & 0.2425 & -- & Best \\
\midrule
\multirow{4}{*}{\textbf{Clusters ($G$)}} 
& 3 & 61 & 0.1718 & 0.0365 & 8  & 15.678 & 0.2425 & -- & Best \\
& 5 & 5  & 0.0370 & 0.0095 & 3  & 3.224  & 0.0113 & -- & Weak \\
& 7 & 11 & 0.0753 & 0.0393 & 8  & 10.044 & 0.0093 & -- & Weak \\
& 9 & 0  & 0.0000 & 0.0406 & 32 & 5.994  & -0.0127 & -- & Unstable \\
\midrule
\multirow{4}{*}{\textbf{$\lambda_{2,1}$ (Sparsity)}} 
& 0.05 & 61 & 0.1718 & 0.0365 & 8 & 15.6777 & 0.2425 & 0.2188 & Strong \\
& 0.10 & 6  & 0.0169 & 0.0019 & 6 & 0.9652  & 0.0130 & 0.2997 & Over-reg. \\
& 0.15 & 0  & 0.0000 & 0.0005 & 6 & 0.9391  & 0.0029 & 0.3015 & Failed \\
& 0.20 & 0  & 0.0000 & 0.0005 & 6 & 0.9391  & 0.0029 & 0.3033 & Failed \\
\bottomrule
\end{tabular}
}
\vspace{-1mm}
\end{table}
$G$ determines how finely respondents are partitioned in latent space. A moderate value produces stable and interpretable group structure, while larger values fragment the population, reducing intervention stability and effectiveness. In particular, Table~\ref{tab:sensitivity_analysis} shows that $G=3$ gives the strongest results, while larger values reduce stability and, in the case of $G=9$, lead to ineffective or unstable behavior.

The sparsity weight $\lambda_{2,1}$ controls the trade-off between intervention strength and intervention simplicity. Lower values allow effective yet compact interventions, whereas higher values overly constrain the solution, suppressing conversion and probability gains. This shows that sparsity is important for interpretability, but excessive regularization weakens the intervention itself.

\subsection{SOTA Analysis}

A direct state-of-the-art (SOTA) comparison for our problem setting is not available, since existing counterfactual explanation methods are primarily designed for instance-level recourse rather than population-level intervention. To establish a principled empirical reference, we select two representative comparison methods aligned with key aspects of our formulation: (i) CEILS~\cite{crupi2024counterfactual}, which incorporates causal structure for instance-level recourse, and (ii) DCE~\cite{you2024distributional}, which performs distribution-level optimization using optimal transport.

CEILS learns a structural causal model from a user-specified graph and generates counterfactuals either in the observed feature space (``Original'') or in a latent residual space (``CEILS''), which are mapped back through structural equations. As an instance-level method, it provides individual recourse actions. We adapt CEILS to our survey setting using a compatible feature subset, a simplified causal graph, and feasibility constraints over mutable and immutable variables. As shown later in Table~\ref{sota_comparison}, both instance-level methods achieve identical conversion performance, indicating that incorporating causal structure alone does not substantially improve behavioral conversion in this setting.
\begin{table}[H]
\centering
\caption{Comparison with representative counterfactual reference methods on the survey dataset.}
\label{sota_comparison}
\resizebox{\columnwidth}{!}{
\begin{tabular}{lccccc}
\toprule
Method & Setting & Eval. Size & $N_{\mathrm{conv}}$ & $R_{\mathrm{conv}}$ & Notes \\
\midrule
Original & Instance & 165 & 27 & 0.1636 & Standard CF (no causality) \\
CEILS & Instance & 165 & 27 & 0.1636 & Causal CF (SCM-based) \\
DCE (OT-based) & Distribution & 33 & 3 & 0.0909 & OT-based CF (test-set subset) \\
\midrule
\textbf{Ours} & Population & 355 & \textbf{61} & \textbf{0.1718} & Sparse OT-based group intervention \\
\bottomrule
\end{tabular}
}
\end{table}
DCE serves as a distribution-level reference because it also adopts an optimal-transport-based objective. However, DCE was originally developed for distributional counterfactual explanation rather than survey-based policy intervention. In our adaptation, we define target and reference groups based on the observed outcome, restrict mutable variables to controllable survey attributes, and enforce feasibility constraints over mixed-type variables. Due to the mismatch between DCE's native formulation and the structured constraints of survey-based intervention, DCE is evaluated on a reduced compatible test-set subset. In addition, the optimization does not always find fully feasible solutions under these constraints and may rely on the best approximate feasible output. Therefore, its results should be interpreted as a practical distributional counterfactual reference rather than a directly matched population-intervention baseline.

As shown in Table~\ref{sota_comparison}, DCE achieves limited conversion ($N_{\mathrm{conv}}=3$, $R_{\mathrm{conv}}=0.0909$). This reflects the gap between distributional counterfactual alignment and the requirement of structured, policy-feasible intervention under mixed-type survey constraints.

\begin{table}[t]
\centering
\caption{Intervention characteristics of the compared reference methods.}
\label{intervention_characteristics}
\resizebox{\columnwidth}{!}{
\begin{tabular}{lccc}
\toprule
Method & Statistic Type & Active Levers & Effort \\
\midrule
Original & Avg. per instance & 3.22 & 2.0367 \\
CEILS & Avg. per instance & 2.85 & 2.0379 \\
DCE (OT-based) & Avg. per batch sample & 0.18 & 0.1818 \\
\midrule
\textbf{Ours} & Group-level intervention & 8 & 15.6777 \\
\bottomrule
\end{tabular}
}
\end{table}

The compared methods operate at different intervention levels, as reflected in Tables~\ref{sota_comparison} and~\ref{intervention_characteristics}. Original and CEILS modify each sample individually, while DCE operates on batches of samples. In contrast, the proposed method learns one coordinated intervention for the full target group.

The lower effort of the reference methods therefore reflects small, localized counterfactual changes rather than a shared population-level policy action. Although the proposed method has higher total effort, it explicitly models target-aware factor importance and shared intervention magnitude through Shapley-guided prioritization and sparse optimal-transport alignment, making the learned intervention more suitable for practical group-level policy design.

\subsection{Generalization Across Domains}

On MNIST, we treat the task as a source-to-target shift problem between two digit classes (3$\rightarrow$8). In this setting, each image is treated as an individual sample, where the source digit class represents the initial group and the target digit class serves as the reference group. The intervention is implemented as a constrained perturbation over editable pixel features, and conversion is measured by whether the modified sample transitions toward the target class under the classifier. The results in Table~\ref{tab:generalization_results} show that the method achieves strong conversion and probability lift in this simplified domain, although the intervention involves a relatively large number of editable features and higher overall effort due to the high-dimensional pixel space. 

\begin{table}[H]
\centering
\caption{Generalization results on MNIST and the VTA survey.}
\label{tab:generalization_results}
\scriptsize
\setlength{\tabcolsep}{2pt}
\resizebox{\columnwidth}{!}{
\begin{tabular}{l l c c c c c c c}
\toprule
Dataset & Task & $N_{\mathrm{conv}}$ & $R_{\mathrm{conv}}$ & Mean $\Delta p$ & Effort $(\|\Delta\|_{2,1})$ & Active Levers & $\Delta W$ & $\rho$ \\
\midrule
\textbf{MNIST (3$\rightarrow$8)} & Image-level shift & 200 & 0.40 & 0.401 & 619.30 & 80 & 4.43 & 0.416 \\
\textbf{VTA Survey} & Behavioral shift & 733 & 0.227 & 0.265 & 155.02 & 4 & 9.03 & 0.057 \\
\bottomrule
\end{tabular}
}
\end{table}

\begin{figure}[H]
    \centering
    \subfloat[Successful conversions (3$\rightarrow$8).]{
        \includegraphics[height=3.2cm]{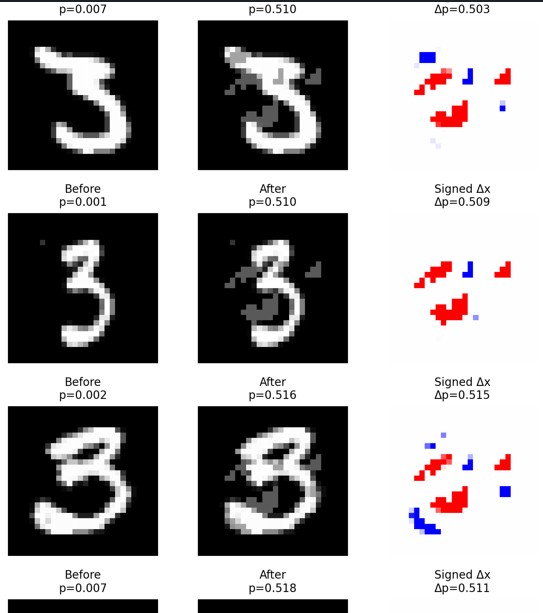}
    }\hfill
    \subfloat[t-SNE visualization of latent-space movement.]{
        \includegraphics[height=3.2cm]{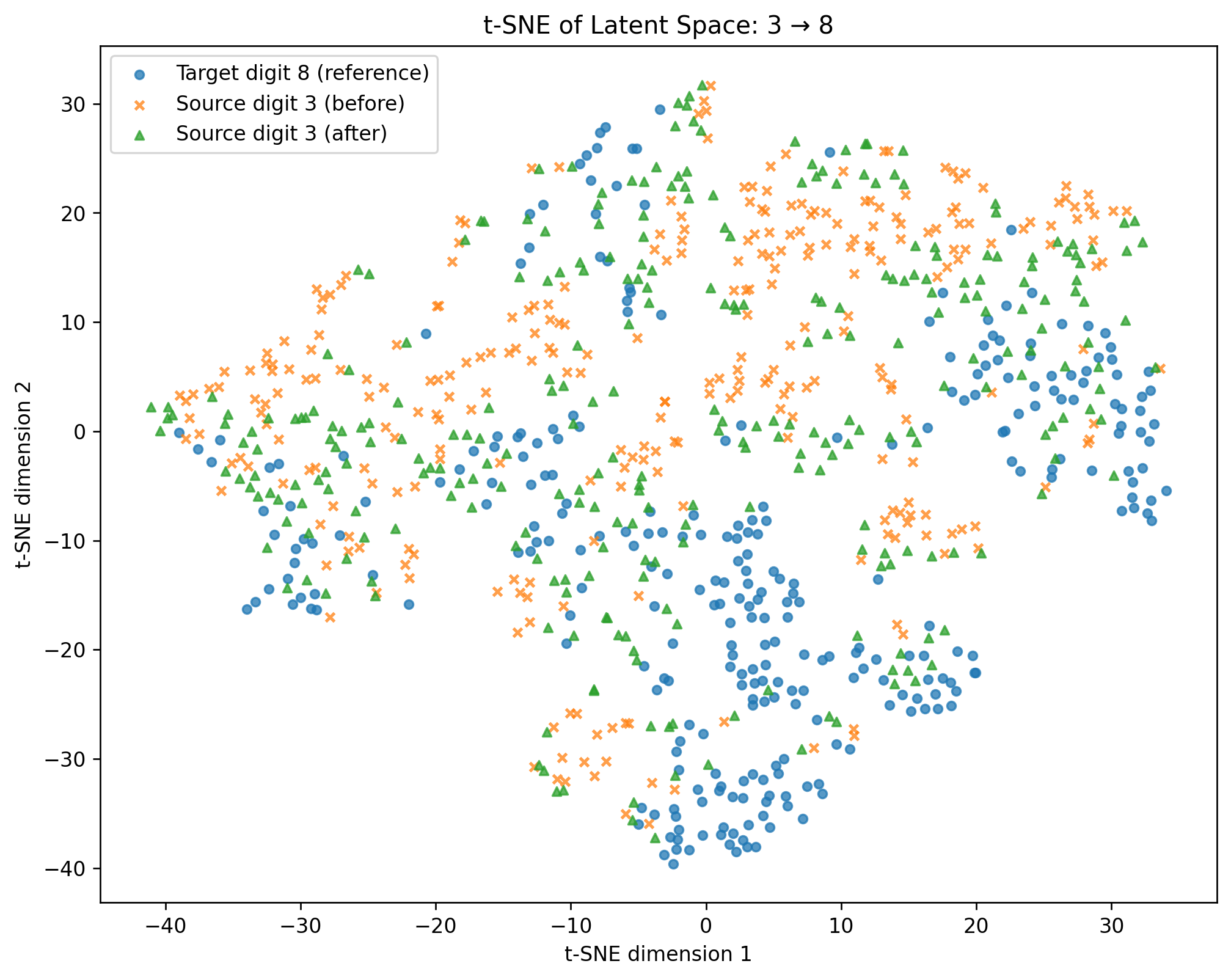}
    }
    \caption{MNIST 3$\rightarrow$8 conversion: representative samples and latent-space movement.}
    \label{fig:mnist_results}
\end{figure}

More importantly, Fig.~\ref{fig:mnist_results}(b) shows that post-intervention samples move in latent space toward the target-digit region, providing geometric evidence of the intended source-to-target transition. Fig.~\ref{fig:mnist_results}(a) illustrates representative successful conversions, where minimal structured perturbations shift digit 3 toward digit 8. Additional qualitative examples, full-grid visualizations, and signed perturbation maps are provided in the Supplementary Material (Fig. 1 \& 2)

For the VTA survey, we formulate the task as a group-to-group conversion problem based on rider satisfaction. Each survey respondent is treated as an individual sample, where users with lower satisfaction (Q7F) form the target group and higher-satisfaction users define the reference group. The intervention operates over a restricted set of service-related controllable variables (e.g., Q7A--Q7E), while demographic and contextual attributes remain fixed. The results in Table~\ref{tab:generalization_results} show that the method achieves consistent conversion and probability improvement, while activating only a small number of policy levers, indicating that sparse and interpretable interventions can be obtained in a real-world survey setting.

\subsection{Characteristics and Interpretation of the Proposed Method}

In addition to aggregate conversion performance, we analyze how the proposed method operates at the group, optimization, and policy levels. Specifically, we examine (i) whether the learned intervention induces coherent movement of the target group toward the reference group in latent space, (ii) whether the optimization procedure produces stable and consistently improving updates, and (iii) whether the resulting intervention can be interpreted as a small set of actionable policy levers. These analyses provide complementary evidence on the effectiveness, stability, and interpretability of the proposed framework.

\subsubsection{Pre--Post Group Movement Analysis}

The proposed method is intended to induce \emph{group-level} movement (target group toward the reference group), not merely to improve individual prediction scores. To test this directly, we perform a pre--post group movement analysis that compares the reference group, the target group before intervention, and the same target group after intervention. Table~\ref{tab:group_movement} reports three complementary quantities for this purpose: the mean surrogate probability, the centroid distance to the reference group in latent space, and the OT discrepancy to the reference distribution.

\begin{table}[H]
\centering
\caption{Pre--post group movement analysis on the Beijing survey.}
\label{tab:group_movement}
\scriptsize
\setlength{\tabcolsep}{2pt}
\resizebox{\columnwidth}{!}{
\begin{tabular}{lcccc}
\toprule
Group & Num. Users & Mean Probability & Dist. to Ref. Centroid & OT Discrepancy to Ref. \\
\midrule
Reference   & 499 & 0.5703 & 0.0000 & 0.0000 \\
Target (Pre)  & 355 & 0.4339 & 0.3735 & 0.7307 \\
Target (Post) & 355 & 0.4704 & 0.2829 & 0.6774 \\
\bottomrule
\end{tabular}
}
\end{table}

This analysis is important because each quantity captures a different aspect of the intended group shift. The mean probability reflects improvement in the modeled outcome, the centroid distance captures geometric movement of the target group toward the reference group in latent space, and the OT discrepancy measures distribution-level alignment while preserving within-group structure.

Table~\ref{tab:group_movement} shows that, after intervention, the target group exhibits a higher mean probability, a smaller centroid distance to the reference group, and a lower OT discrepancy. Thus, the learned intervention does not simply increase predicted adoption in isolation; it also shifts the target group closer to the reference group in the latent representation and improves alignment at the distributional level.

\subsubsection{Optimization Behavior}

\begin{figure}[H]
    \centering
    \includegraphics[width=\columnwidth]{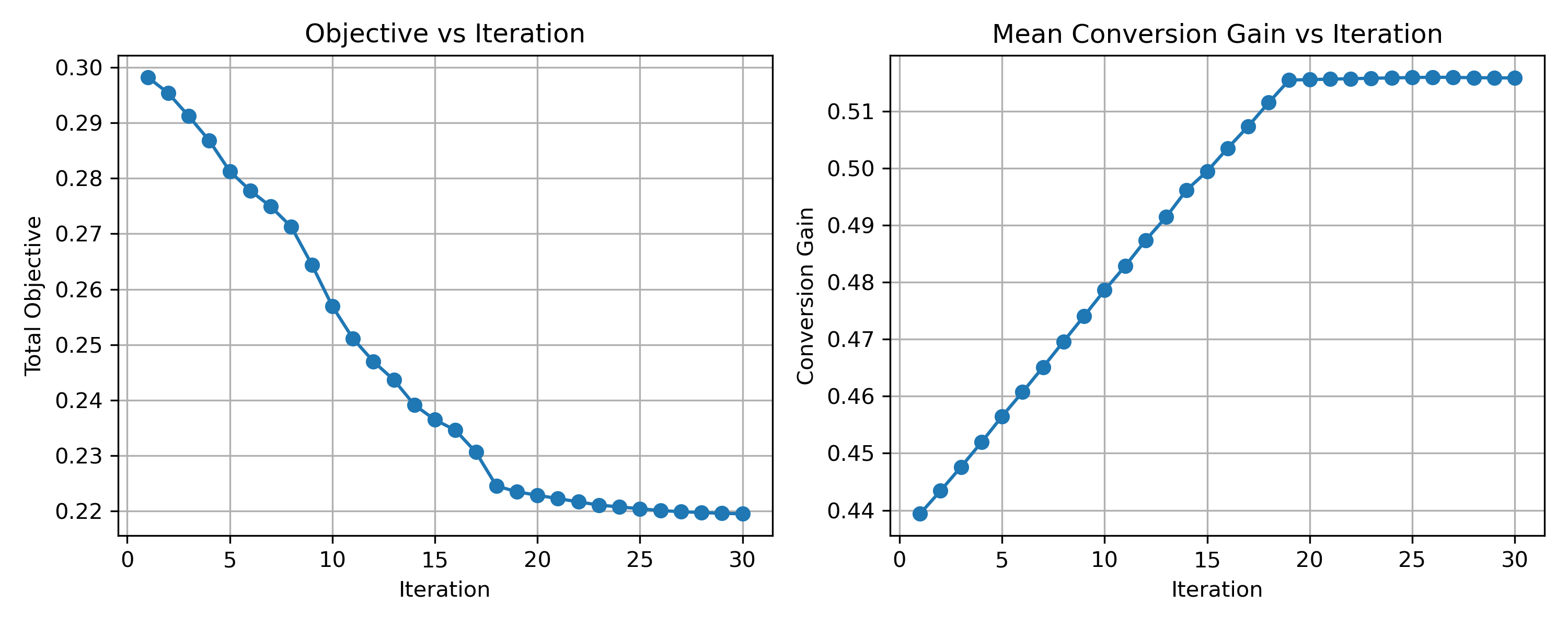}
    \caption{Optimization trajectory showing monotonic objective decrease and saturation of mean conversion gain across iterations.}
    \label{fig:optimization_trajectory}
\end{figure}

To assess the intervention optimizer, we examine the optimization trajectory over accepted iterations. The left panel of Fig.~\ref{fig:optimization_trajectory} shows the total objective, which decreases monotonically and stabilizes after approximately 20 iterations, indicating stable convergence without oscillation. The right panel shows the mean surrogate gain, which increases steadily in the early and middle stages and then plateaus, indicating that the learned intervention becomes progressively more effective before reaching saturation.

Taken together, these trends indicate that the optimization process is stable and well directed: accepted updates consistently improve the objective, while intervention effectiveness increases until no substantial additional gain is obtained. The late-stage plateau in mean surrogate gain suggests that later iterations mainly refine the objective rather than produce large further changes in target-group conversion, supporting convergence to a practically stable solution.

\subsubsection{Policy-Level Interpretation of the Learned Intervention}

A key goal of the proposed framework is not only to improve conversion, but also to identify which controllable policy levers contribute to moving the target group toward the reference group. To make this explicit, Table~\ref{tab:policy_levers_en} reports the active levers selected by the full model, together with their intervention magnitudes and importance weights.

\begin{table}[t]
\centering
\caption{Top activated policy levers on the survey dataset}
\label{tab:policy_levers_en}
\scriptsize
\setlength{\tabcolsep}{2pt}
\resizebox{\columnwidth}{!}{
\begin{tabular}{c l c c}
\toprule
Feature Index & Policy Lever & $\Delta$ (Magnitude) & Importance ($\omega$) \\
\midrule
7  & Exchange for discounted health insurance enrollment & 0.311 & 0.074 \\
5  & Exchange for fitness course coupons or trial cards & 0.261 & 0.076 \\
6  & Exchange for health check-up packages & 0.218 & 0.075 \\
8  & Exchange for personal annual carbon reduction certificate & 0.009 & 0.069 \\
9  & Exchange for tree planting / ecological protection projects & 0.009 & 0.068 \\
10 & Exchange for donations to public welfare projects & 0.008 & 0.061 \\
13 & Group-based low-carbon travel to earn more carbon credits & 0.008 & 0.054 \\
4  & Exchange for transportation benefits (bus/metro/shared bike) & 0.008 & 0.042 \\
\bottomrule
\end{tabular}
}
\end{table}

The table shows that the learned intervention is concentrated on a small and interpretable subset of variables rather than being diffused across many features. It also links the final intervention to the latent-to-feature prioritization step, since the most strongly adjusted levers are also assigned relatively high importance weights. This indicates that the optimization is not selecting variables arbitrarily, but is acting consistently on the controllable levers identified as most influential for the modeled outcome. As a result, the learned intervention remains interpretable as a focused set of actionable policy adjustments rather than an opaque high-dimensional modification.

\section{Conclusion}
We studied sparse counterfactual community intervention from survey responses, where the objective is to shift a target group toward a desired reference group through feasible adjustments on controllable survey variables. We formulated this task as a sparse distributional alignment problem under a fixed-coordinate latent representation. To address this problem, we proposed a conversion-by-alignment framework that combines fixed-basis latent representation, target-aware latent-to-feature prioritization, and entropy-regularized optimal-transport alignment with weighted $\ell_{2,1}$ sparsity. The framework learns compact and interpretable group-level interventions while preserving feasibility and pre/post latent comparability. Experiments on real-world transportation surveys demonstrated consistent improvements in target-group conversion and latent-space alignment with sparse policy adjustments. The proposed framework provides a general direction for transforming survey analysis from descriptive modeling toward actionable group-level intervention design.

\begingroup
\small 
\bibliographystyle{IEEEtran}
\bibliography{ref}

@article{borhan2014predicting,
  title={Predicting the use of public transportation: a case study from Putrajaya, Malaysia},
  author={Borhan, Muhamad Nazri and Syamsunur, Deprizon and Mohd Akhir, Norliza and Mat Yazid, Muhamad Razuhanafi and Ismail, Amiruddin and Rahmat, Riza Atiq},
  journal={The Scientific World Journal},
  volume={2014},
  number={1},
  pages={784145},
  year={2014},
  publisher={Wiley Online Library}
}

@article{liu2025incentives,
  title={How incentives affect commuter willingness for public transport: Analysis of travel mode shift across various cities},
  author={Liu, Bing and Ma, Zhenliang and Kong, Hui and Ma, Xiaolei},
  journal={Travel Behaviour and Society},
  volume={39},
  pages={100966},
  year={2025},
  publisher={Elsevier}
}

@article{jing2022impact,
  title={The impact of public transportation on carbon emissions—from the perspective of energy consumption},
  author={Jing, Qin-Lei and Liu, Han-Zhen and Yu, Wei-Qing and He, Xu},
  journal={Sustainability},
  volume={14},
  number={10},
  pages={6248},
  year={2022},
  publisher={MDPI}
}

@article{yeganeh2018social,
  title={A social equity analysis of the US public transportation system based on job accessibility},
  author={Yeganeh, Armin Jeddi and Hall, Ralph P and Pearce, Annie R and Hankey, Steve},
  journal={Journal of Transport and Land Use},
  volume={11},
  number={1},
  pages={1039--1056},
  year={2018},
  publisher={JSTOR}
}

@misc{antipova2020accessibility,
  title={Accessibility and transportation equity},
  author={Antipova, Anzhelika and Sultana, Salima and Hu, Yujie and Rhudy Jr, James P},
  journal={Sustainability},
  volume={12},
  number={9},
  pages={3611},
  year={2020},
  publisher={MDPI}
}

@article{wang2025counterfactual,
  title={Counterfactual explanations for deep learning-based traffic forecasting},
  author={Wang, Rushan and Xin, Yanan and Zhang, Yatao and Perez-Cruz, Fernando and Raubal, Martin},
  journal={Communications in Transportation Research},
  volume={5},
  pages={100176},
  year={2025},
  publisher={Elsevier}
}

@article{you2024distributional,
  title={Distributional counterfactual explanations with optimal transport},
  author={You, Lei and Cao, Lele and Nilsson, Mattias and Zhao, Bo and Lei, Lei},
  journal={arXiv preprint arXiv:2401.13112},
  year={2024}
}

@article{kim2025using,
  title={Using a Technology Acceptance Model to explore the intention to use digital health technologies among people with disabilities: cross-sectional survey study},
  author={Kim, Jae-Hak and Kim, Janghyeon and Youn, Bo-Young},
  journal={Journal of Medical Internet Research},
  volume={27},
  pages={e79595},
  year={2025},
  publisher={JMIR Publications Toronto, Canada}
}

@article{susanto2025investigating,
  title={Investigating consumers’ behavioral intentions in the adoption of 5G mobile networks: a holistic approach to technology acceptance and business process integration},
  author={Susanto, Heru and Hj Ahamad, Izzati Nadiah and Shafa Susanto, Alifya Kayla},
  journal={Frontiers in Communications and Networks},
  volume={6},
  pages={1594378},
  year={2025},
  publisher={Frontiers Media SA}
}

@article{srivastava2011case,
  title={A case study and survey-based assessment of the management of innovation and technology},
  author={Srivastava, Mukesh},
  journal={Journal of technology management \& innovation},
  volume={6},
  number={1},
  pages={147--160},
  year={2011},
  publisher={SciELO Chile}
}

@article{zhang2024promoting,
  title={Promoting green transportation through changing behaviors with low-carbon-travel function of digital maps},
  author={Zhang, Li and Tao, Lan and Yang, Fangyi and Bao, Yuchen and Li, Chong},
  journal={Humanities and Social Sciences Communications},
  volume={11},
  number={1},
  pages={1--10},
  year={2024},
  publisher={Palgrave}
}

@article{de2023possible,
  title={Is it possible to attract private vehicle users towards public transport? Understanding the key role of service quality, satisfaction and involvement on behavioral intentions},
  author={de O{\~n}a, Juan and de O{\~n}a, Roc{\'\i}o},
  journal={Transportation},
  volume={50},
  number={3},
  pages={1073--1101},
  year={2023},
  publisher={Springer}
}

@article{rong2022impact,
  title={Impact analysis of actual traveling performance on bus passenger’s perception and satisfaction},
  author={Rong, Rui and Liu, Lishan and Jia, Ning and Ma, Shoufeng},
  journal={Transportation Research Part A: Policy and Practice},
  volume={160},
  pages={80--100},
  year={2022},
  publisher={Elsevier}
}

@article{ye2025private,
  title={Private car users’ willingness to switch to public transportation and its influencing factors in the Yangtze River Delta},
  author={Ye, Xiaohui and Sato, Masayuki},
  journal={Asian Transport Studies},
  volume={11},
  pages={100171},
  year={2025},
  publisher={Elsevier}
}

@article{sogbe2025scaling,
  title={Scaling up public transport usage: a systematic literature review of service quality, satisfaction and attitude towards bus transport systems in developing countries},
  author={Sogbe, Eugene and Susilawati, Susilawati and Pin, Tan Chee},
  journal={Public Transport},
  volume={17},
  number={1},
  pages={1--44},
  year={2025},
  publisher={Springer}
}

@article{ashraf2025importance,
  title={Importance-aware Topic Modeling for Discovering Public Transit Risk from Noisy Social Media},
  author={Ashraf, Fatima and Sabir, Muhammad Ayub and Deng, Jiaxin and Pang, Junbiao and Yu, Haitao},
  journal={arXiv preprint arXiv:2512.06293},
  year={2025}
}

@article{li2025topic,
  title={Topic modeling help enhancing sustainable mobility},
  author={Li, Xiao and He, Guangxi and Guo, Peng and Guo, Zhaohua and Lin, Shicheng and Du, Shouji},
  journal={Journal of Cleaner Production},
  volume={534},
  pages={147068},
  year={2025},
  publisher={Elsevier}
}

@article{yang2023identifying,
  title={Identifying latent activity behaviors and lifestyles using mobility data to describe urban dynamics},
  author={Yang, Yanni and Pentland, Alex and Moro, Esteban},
  journal={EPJ Data Science},
  volume={12},
  number={1},
  pages={15},
  year={2023},
  publisher={Springer Berlin Heidelberg}
}

@article{aminpour2025unveiling,
  title={Unveiling mobility patterns beyond home/work activities: A topic modeling approach using transit smart card and land-use data},
  author={Aminpour, Nima and Saidi, Saeid},
  journal={Travel Behaviour and Society},
  volume={38},
  pages={100905},
  year={2025},
  publisher={Elsevier}
}

@article{kriswardhana2025uncovering,
  title={Uncovering distinct public transport user profiles and the factors influencing the users’ intentions},
  author={Kriswardhana, Willy and Ismael, Karzan and Duleba, Szabolcs and Eszterg{\'a}r-Kiss, Domokos},
  journal={Journal of Urban Mobility},
  volume={7},
  pages={100127},
  year={2025},
  publisher={Elsevier}
}

@article{na2023toward,
  title={Toward practical and plausible counterfactual explanation through latent adjustment in disentangled space},
  author={Na, Seung-Hyup and Nam, Woo-Jeoung and Lee, Seong-Whan},
  journal={Expert Systems with Applications},
  volume={233},
  pages={120982},
  year={2023},
  publisher={Elsevier}
}

@article{crupi2024counterfactual,
  title={Counterfactual explanations as interventions in latent space},
  author={Crupi, Riccardo and Castelnovo, Alessandro and Regoli, Daniele and San Miguel Gonzalez, Beatriz},
  journal={Data Mining and Knowledge Discovery},
  volume={38},
  number={5},
  pages={2733--2769},
  year={2024},
  publisher={Springer}
}

@article{kim2024cirf,
  title={CIRF: Importance of related features for plausible counterfactual explanations},
  author={Kim, Hee-Dong and Ju, Yeong-Joon and Hong, Jung-Ho and Lee, Seong-Whan},
  journal={Information Sciences},
  volume={678},
  pages={120974},
  year={2024},
  publisher={Elsevier}
}

@misc{vta2013,
  title = {Santa Clara Valley On-Board Transit Survey (2013)},
  howpublished = {\url{https://www.nlr.gov/transportation/secure-transportation-data/tsdc-santa-clara-valley-onboard-transit-survey}},
  year = {2013}
}

@article{saberi2025nonnegative,
  title={Nonnegative matrix factorization in dimensionality reduction: A survey},
  author={Saberi-Movahed, Farid and Berahmand, Kamal and Sheikhpour, Razieh and Li, Yuefeng and Pan, Shirui and Jalili, Mahdi},
  journal={ACM Computing Surveys},
  volume={58},
  number={5},
  pages={1--41},
  year={2025},
  publisher={ACM New York, NY}
}

@article{borgonovo2024many,
  title={The many Shapley values for explainable artificial intelligence: A sensitivity analysis perspective},
  author={Borgonovo, Emanuele and Plischke, Elmar and Rabitti, Giovanni},
  journal={European Journal of Operational Research},
  volume={318},
  number={3},
  pages={911--926},
  year={2024},
  publisher={Elsevier}
}

@article{verma2024counterfactual,
  title={Counterfactual explanations and algorithmic recourses for machine learning: A review},
  author={Verma, Sahil and Boonsanong, Varich and Hoang, Minh and Hines, Keegan and Dickerson, John and Shah, Chirag},
  journal={ACM Computing Surveys},
  volume={56},
  number={12},
  pages={1--42},
  year={2024},
  publisher={ACM New York, NY}
}

@book{peyre2019computational,
  title={Computational optimal transport: With applications to data science},
  author={Peyr{\'e}, Gabriel and Cuturi, Marco},
  year={2019},
  publisher={Now Foundations and Trends}
}
\endgroup

\end{document}